\documentclass[letterpaper, 10 pt, conference]{ieeeconf}  

\IEEEoverridecommandlockouts                              
\usepackage[most]{tcolorbox} 
\newtcolorbox{mainbox}[2][]{%
  enhanced,
  skin=bicolor,            
  breakable,               
  title=#2,
  fonttitle=\bfseries,
  colframe=black!75,
  colback=black!5,
  colbacklower=black!10,   
  fontlower=\itshape,      
  arc=3pt, boxrule=0.5pt,
  left=5pt, right=5pt, top=5pt, bottom=5pt,
  before skip=4pt plus 2pt minus 1pt,
  after  skip=4pt plus 2pt minus 1pt,
  #1
}
\newtcolorbox{nestedbox}[1]{%
  enhanced, breakable,
  colback=white, colframe=black!60,
  boxrule=0.5pt, arc=2pt, title=#1
}

\usepackage{graphicx}
\usepackage{booktabs}
\usepackage{amsmath}
\usepackage{siunitx}
\usepackage{caption}
\usepackage{subcaption}
\usepackage[hidelinks]{hyperref}

\title{\LARGE \bf\texttt{GUIDES}: Guidance Using Instructor-Distilled Embeddings for\\ Pre-Trained Robot Policy Enhancement}

\author{Minquan Gao$^{1,2*}$, Xinyi Li$^{2*}$, Qing Yan$^{2*}$, Xiaojian Sun$^{2}$, Xiaopan Zhang$^{1}$, Chien-Ming Huang$^{2\ddag}$, Jiachen Li$^{1\ddag}$  
\thanks{$^{*}$ Equal contribution}
\thanks{$^{\ddag}$ Corresponding authors}
\thanks{$^{1}$ University of California, Riverside. \{xzhan006, jiachen.li\}@ucr.edu}
\thanks{$^{2}$ Johns Hopkins University. \{mgao40, xli383, qyan13, xsun90\}@jh.edu, cmhuang@cs.jhu.edu}
}
\begin{document}

\maketitle
\thispagestyle{empty}
\pagestyle{empty}
\begin{abstract} 
Pre-trained robot policies serve as the foundation of many validated robotic systems, which encapsulate extensive embodied knowledge. However, they often lack the semantic awareness characteristic of foundation models, and replacing them entirely is impractical in many situations due to high costs and the loss of accumulated knowledge. 
To address this gap, we introduce \texttt{GUIDES}, a lightweight framework that augments pre-trained policies with semantic guidance from foundation models without requiring architectural redesign. 
\texttt{GUIDES} employs a fine-tuned vision-language model (Instructor) to generate contextual instructions, which are encoded by an auxiliary module into guidance embeddings. These embeddings are injected into the policy’s latent space, allowing the legacy model to adapt to this new semantic input through brief, targeted fine-tuning. 
For inference-time robustness, a large language model–based Reflector monitors the Instructor’s confidence and, when confidence is low, initiates a reasoning loop that analyzes execution history, retrieves relevant examples, and augments the VLM’s context to refine subsequent actions. 
Extensive validation in the RoboCasa simulation environment across diverse policy architectures shows consistent and substantial improvements in task success rates. 
Real-world deployment on a UR5 robot further demonstrates that \texttt{GUIDES} enhances motion precision for critical sub-tasks such as grasping. Overall, \texttt{GUIDES} offers a practical and resource-efficient pathway to upgrade, rather than replace, validated robot policies.
\end{abstract}

\section{Introduction} \label{sec:intro}

\begin{figure*}[!t]
    \centering
     \includegraphics[width=\linewidth]{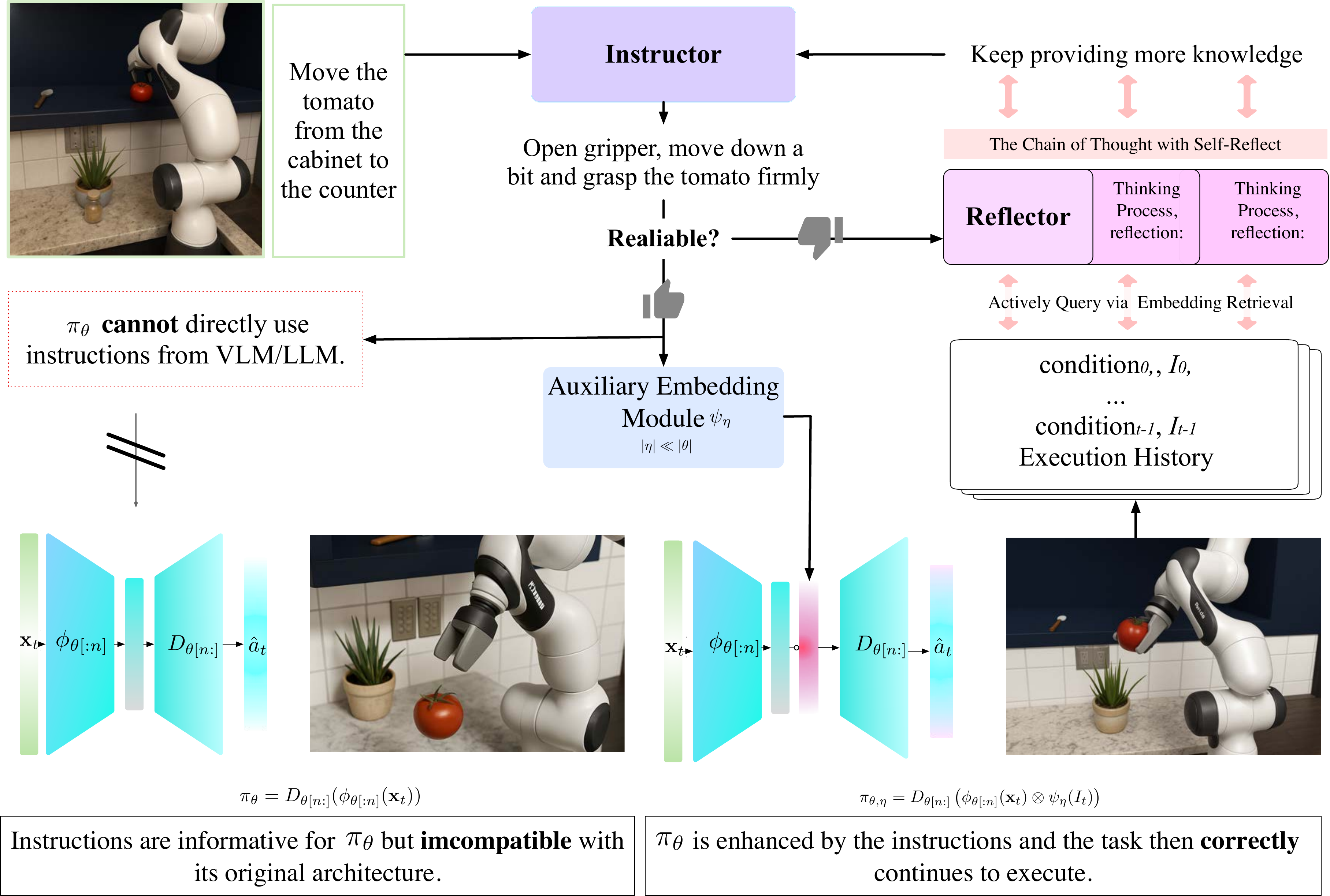}
    \caption{Overview of the \texttt{Guides} framework. The \textsc{Instructor}, initially fine-tuned with motion ground truth, provides step-wise instructions. These are mapped to guidance embedding via an auxiliary module, and integrated with the original model’s latent space $\phi$ to enhance $\pi_\theta$. Meanwhile, the \textsc{Reflector} uses Chain-of-Thought reasoning to analyze execution and infer potential risks or next steps. It queries via embedding retrieval based on the execution history.}
    \label{fig:arch}
\end{figure*}

Integrating foundation models, including vision-language Models (VLMs)~\cite{zhang2024vision,ghosh2024exploring,wang2024qwen2,team2023gemini,nag2025conformal,fan2026vlm} and large language models (LLMs)~\cite{touvron2023llama,naveed2023comprehensive}, into robot policies has driven significant advances in high-level semantic reasoning and task generalization~\cite{li2023vision,zawalski2024robotic,li2024improving,Kim2024,yan2025rdd,wang2026drive,zhang2025lamma,zhang2026commcp,chakraborty2025heal,gong2026autofocus,kim2025human}. End-to-end architectures, such as RT-2~\cite{Brohan2023} and OpenVLA~\cite{Kim2024}, demonstrate remarkable emergent capabilities through training on massive multi-modal datasets.
While foundation models represent a promising direction for next-generation robotics, they also introduce a fundamental challenge for existing systems. Many deployed robots still rely on traditional policies, such as BC-Transformer~\cite{mandlekar2021what} or Diffusion Policy~\cite{Chi2023Diffusion}, which, although lacking semantic awareness, have been extensively trained and validated on real-world data, making them reliable and valuable. Replacing these systems with monolithic VLA architectures risks discarding accumulated knowledge and would incur prohibitive re-validation costs and deployment risks.
Therefore, in this work, we propose a distinct design regime: \textit{augmenting existing validated policies with compact semantic guidance rather than replacing them entirely}.

This leads to a key research question: How can we design a semantic co-processor that seamlessly integrates with diverse pre-trained policies to empower them with foundation model capabilities without modifying their core, validated architectures?
To address this, we introduce \texttt{GUIDES} (Guidance Using Instructor-Distilled Embeddings), a lightweight and architecture-agnostic framework that provides a principled mechanism for integrating foundation model capabilities into pre-trained robot policies, as illustrated in Figure~\ref{fig:arch}.
A fine-tuned VLM, termed the \textsc{Instructor}, generates contextual action instructions that are distilled by a lightweight auxiliary module into compact guidance embeddings, which are injected directly into the policy’s latent space.
Crucially, \texttt{GUIDES} enhances decision making robustness beyond simple instruction-following through a novel \textsc{Reflector} module at inference time. 
The \textsc{Reflector} forms a dynamic online reasoning loop: it employs an LLM to analyze execution history, diagnose potential failures using Chain-of-Thought reasoning~\cite{zawalski2024robotic}, and retrieve relevant knowledge from past experiences. This reflective mechanism significantly improves performance in long-horizon, high-complexity tasks where subtle errors may otherwise cascade into mission failure.
Integrating \texttt{GUIDES} requires no architectural modification to the base policy. A brief fine-tuning stage is sufficient to train the auxiliary module and minimally adapt the original policy parameters to the new semantic inputs. This non-invasive design preserves the integrity of the pre-trained architecture while enabling compatibility with VLM and LLM tools. 

The main contributions of this work are as follows:
\begin{itemize}
\item We propose \texttt{GUIDES}, an architecture-agnostic framework that augments existing, validated robot policies with foundation model capabilities, preserving original architectures and leveraging pre-trained knowledge.

\item We introduce a novel inference-time reflection mechanism that leverages LLM-based Chain-of-Thought reasoning to analyze execution history, anticipate failures, and improve task success rates.
\item We validate \texttt{GUIDES} across heterogeneous policy architectures (e.g., Transformers, Diffusion Models) on the RoboCasa benchmark~\cite{Nasiriany2024RoboCasa} and demonstrate successful real-world deployment on a UR5 robot, which shows consistent performance gains and strong cross-architecture generalization.
\end{itemize}

\section{Related Work}

\subsection{End-to-End Foundation Model Integration}

A growing body of work focuses on directly integrating large foundation models into robotic policies~\cite{da2024prompt}. Unified architectures like RT-2~\cite{Brohan2023} and OpenVLA~\cite{Kim2024} combine visual perception, language understanding, and control in an end-to-end fashion. Similarly, language-conditioned methods, such as Perceiver-Actor~\cite{shridhar2023perceiver} and RACER~\cite{Dai2024}, incorporate textual inputs directly into the policy network. While these approaches demonstrate impressive capabilities, they require significant architectural changes and extensive training, which hinders the reuse of existing pre-trained models. 
In contrast, we demonstrate how foundation model guidance can be effectively incorporated while preserving the architecture and pre-trained weights of existing policies.

\subsection{Efficient and Decoupled VLA Architectures}

To address the computational demands of monolithic models, several studies explore more efficient or modular architectures. For instance, TinyVLA~\cite{wen2024tinyvla} constructs compact end-to-end policies by initializing from smaller pre-trained models and integrating components such as a diffusion policy decoder. RoboFlamingo~\cite{li2023vision}, on the other hand, adopts a decoupled design, using a frozen VLM to extract visual-language features passed to a separate trainable policy head. While these methods improve efficiency, they still require a specific architectural redesign, either by adopting a compact model from scratch or constructing a composite VLM-plus-head framework. This makes them less suitable for retrofitting existing policies without extensive modification. In contrast, \texttt{GUIDES} enhances legacy policies in a minimally invasive manner.

\subsection{Hierarchical Task Planning Integration}

Another line of research uses VLMs for high-level task planning instead of direct control. Hierarchical methods, such as VLM-TAMP~\cite{Yang2024VLMTAMP}, interpret natural language instructions using a VLM to generate symbolic subgoals, which are then passed to a Task and Motion Planning (TAMP) system. While this leverages the reasoning strength of VLMs, it enforces a rigid two-level control hierarchy. Such approaches are incompatible with monolithic end-to-end policies that lack a built-in hierarchical structure. \texttt{GUIDES}, however, directly augments a single policy architecture without requiring an explicit task planner, which preserves the flexibility and modularity of existing models.

\subsection{Behavior Cloning}

Behavior Cloning (BC)~\cite{torabi2018behavioral,florence2022implicit} is an imitation learning approach that trains policies by mimicking expert demonstrations~\cite{Pomerleau1989ALVINN,abbeel2004apprenticeship,argall2009survey}.
However, BC often struggles in real-world applications due to sensitivity to small deviations from optimal behavior. In Figure~\ref{fig:arch}, for the task ``pick up the tomato and place it on the counter'', even a slight misalignment between the gripper and the tomato can lead to failure, regardless of how optimal the remaining trajectory is. \texttt{GUIDES} introduces auxiliary semantic guidance at each time step based on the current state. Through post-training, both the original policy and the auxiliary module learn to map these semantic cues to improved action selection, enhancing performance step by step.

In summary, we position our approach as an ``upgrade'' route rather than a ``replacement''. End-to-end VLAs (e.g., RT-2, OpenVLA) restructure the perception–action stack and rely on large, cross-embodiment corpora; modular designs (e.g., RACER, TinyVLA/RoboFlamingo) still introduce new conditioning pathways or heads. By instead fusing compact guidance into the incumbent latent space, we avoid interface changes and re-qualification overheads, making our method complementary to VLAs rather than a competing full replacement \cite{Brohan2023,Kim2024,Dai2024,wen2024tinyvla,li2023vision}.
\section{Methods} \label{sec:method}
We formulate our task as a behavior cloning (BC) problem. The objective is to learn a policy $\pi_\theta$ that maps observations $x_t$ to actions $a_t$ by imitating expert demonstrations $\{(x_t, a_t)\}_{t=1}^N$. At each time step $t$, the agent observes $x_t$, and the expert provides the corresponding action $a_t$. The policy is trained to maximize the likelihood of expert actions conditioned on the observed states:
$\mathcal{L}_{\text{BC}}(\theta) = - \frac{1}{N} \sum_{t=1}^{N} \log \pi_\theta(a_t \mid x_t),$
where $\theta$ denotes the parameters of the policy. Performance is evaluated using the task success rate in test environments, where a task is considered successful if completed within a fixed horizon of $H$ steps.
\texttt{GUIDES} enhances a base robot policy $\pi_\theta$ with semantic knowledge from foundation models, without modifying the underlying architecture. The framework introduces three key components: (1) an \textsc{Instructor}, a fine-tuned VLM that generates step-wise textual instructions; (2) an auxiliary embedding module $\psi_\eta$ that encodes these instructions into compact guidance embeddings $\mathbf{g}_t$, which are injected into the policy’s latent space; and (3) a \textsc{Reflector} module, comprising an LLM and VLM, which operates at inference time to refine instructions via chain-of-thought reasoning. Figure~\ref{fig:arch} illustrates the complete system pipeline.
\paragraph{\textsc{Instructor}} 

The \textsc{Instructor}, powered by a vision-language model (VLM), generates step-wise action instructions $I_{str}^t$. A key challenge is that general-purpose VLMs often struggle with the fine-grained spatial reasoning required for robotic manipulation~\cite{hudson2019gqa,johnson2017clevr}. To address this, we fine-tune the \textsc{Instructor} using a novel two-stage procedure that distills spatial knowledge from expert demonstrations: \textit{1) Ground-Truth Instruction Generation}, and \textit{2) Model Fine-Tuning}.

\textbf{1) Ground-Truth Instruction Generation.}  
We construct a high-quality dataset of text instructions grounded in physical motion. For each time step $t$ in an expert trajectory, we extract the ground-truth motion delta of the end-effector, $\delta M^t$, in the world frame. This delta encodes the precise physical action and is defined as:
\[
\delta M^{t} = \left[
T^{(\mathcal{W}, t+1)} - T^{(\mathcal{W}, t)},
R^{(\mathcal{W}, t+1)} - R^{(\mathcal{W}, t)}
\right].
\]
We then prompt the VLM with the current observation (RGB image and robot state) and the corresponding ground-truth motion delta $\delta M^t$, instructing it to generate a textual description that reflects the given motion. For instance, if $\delta M^t$ corresponds to vertical upward movement, the VLM may produce an instruction like ``lift the tomato vertically out of the bin.'' This yields a set of kinematically grounded textual labels, denoted as $\{I_{str}^{t*}\}$.

\textbf{2) Model Fine-Tuning.}  
We fine-tune the \textsc{Instructor} using these generated labels. During the training phase, the input to the model is a standard prompt $\mathcal{P}_{str}^t$, which includes the task description, RGB observation, and robot state, \emph{but critically excludes} the ground-truth motion $\delta M^t$. The model is trained to predict the corresponding instruction $I_{str}^{t*}$ as the target output. This prompt format is identical to the one used during inference, which is provided in the following.

Through this two-stage process, the \textsc{Instructor} learns to map visual and proprioceptive inputs directly to spatially grounded instructions, effectively internalizing the expert's spatial reasoning from kinematic data without requiring access to such data during inference.

\begin{mainbox}[colbacklower=black!10, fontlower=\itshape]{Prompt for \textsc{Instructor} (Training \& Inference)}
\small
\textbf{Task:} $<$e.g., Pick up the tomato and place it on the counter$>$ \\
\textbf{RGB Observation:} $<$Image data$>$ \\
\textbf{Robot State:}  $<$\quad EEF Position and Gripper state$>$ \\

\textbf{System Instruction for VLM:} You are a helpful robot assistant. Based on the inputs, generate a two-part response describing the current situation and the best next action.
\begin{enumerate}
    \item \textbf{Condition:} Analyze the image and robot state to concisely describe only the key observations critical to task execution or potential hazards.
    \item \textbf{Next Action:} State the immediate next action the robot should perform.
\end{enumerate}

\textbf{Example VLM Output:} \\
\quad \textbf{Condition:} \textit{The gripper is open and positioned directly above the red tomato in the cabinet.} \\
\quad \textbf{Next Action:} \textit{Lower the end-effector vertically to grasp the tomato.}
\end{mainbox}
\paragraph{Auxiliary Encoding Module (AEM)} 

To inject high-level semantic guidance into the policy, we introduce a lightweight auxiliary embedding module $\psi_\eta$. At each time step $t$, it receives a task specification $\mathcal{S}_t = (\mathcal{T}_{str}^k, I_{str}^t)$, which consists of the task description and the step-wise instruction from the \textsc{Instructor} and generates an embedding $\mathbf{g}_t = \psi_\eta(\mathcal{S}_t)$. This embedding is combined with the policy’s latent representation via an integration operator $\otimes$, forming the input to the decoder:
\[
\pi_{\theta, \eta}(a_t \mid x_t) = D_{\theta[n:]}(\phi_{\theta[:n]}(x_t) \otimes \mathbf{g}_t).
\]
We inject $\mathbf{g}_t$ at the encoder output $\phi_{\theta[:n]}$ for two reasons: (1) it shortens the backpropagation path for $\eta$, which minimizes interference with pretrained weights; and (2) it generalizes across architectures with minimal modification.
In our implementation, $\psi_\eta$ employs a cross-attention mechanism to dynamically contextualize the step-wise instruction embedding with the global task. The resulting informed guidance vector $\mathbf{g}_t$ is then processed by a lightweight feed-forward network. To integrate $\mathbf{g}_t$ with the latent feature $\phi_{\theta[:n]}(x_t)$, we adopt element-wise addition, as it introduces negligible computational overhead and preserves the architectural integrity of the policy backbone by avoiding any changes to tensor shapes, ensuring maximum adaptability.
\paragraph{Guidance Awareness Training}

This stage efficiently fine-tunes the policy head ($D_{\theta[n:]}$) and the lightweight guidance branch ($\psi_\eta$), while keeping the large visual encoder ($\phi_{\theta[:n]}$) frozen. To ensure the training is both practical and principled, we address two key challenges: the high computational cost of inference and the need for a sample-efficient fine-tuning schedule.

\textbf{1) Reducing Inference Cost.}  
Querying the \textsc{Instructor} VLM at every step of a long-horizon task (300–1000 steps) is prohibitively expensive. To mitigate this, we reuse the previous instruction when the visual scene remains largely unchanged. 
Specifically, we compute the CLIP cosine similarity between consecutive frames ($s_t = \text{sim}(x_t, x_{t-1})$), and only query the VLM when $s_t \le \tau_{\text{sim}}$ ($\tau_{\text{sim}} = 0.95$). This simple yet effective heuristic reduces wall-clock training time by a factor of 4–10 times.

\textbf{2) Sample-Efficient Fine-Tuning Schedule.}  
Determining an efficient fine-tuning duration without extensive trial-and-error is challenging. To address this, we draw inspiration from PAC learning theory~\cite{anthony2009neural}, which states that the number of training samples ($m$) required scales with the number of learnable parameters ($|\Theta|$). Given that our guidance branch is significantly smaller than the base model ($|\eta| \ll |\theta|$, approximately 0.8\% in our setup), we apply a proportional scaling rule to define a lightweight fine-tuning schedule:
$E_\eta = \frac{|\eta|}{|\theta|}\,E,$
where $E$ is the full training schedule and $E_\eta$ is the schedule for the guidance branch. This approach reduces training epochs to only 5–10\% of the original schedule, ensuring both theoretical soundness and practical efficiency.

\textbf{3) Optimization Objective.}  
The policy head and guidance branch are jointly updated via a guidance-aware behavior cloning loss. Instead of conditioning the policy solely on visual features, we inject semantic guidance by forming a hybrid latent representation, summing the visual features with the language-based guidance embedding:
\[
\mathcal{L}(\theta,\eta)=
-\!\sum_{t=1}^{N}\!
\log\pi_\theta\!\bigl(a_t\mid
\underbrace{\phi(x_t)}_{\text{Visual Features}} +
\underbrace{\psi_\eta(\mathcal{G}_t^{\mathrm{aux}},\mathcal{T}^{\text{cur}})}_{\text{Guidance Embedding}} \bigr).
\]
Conceptually, the guidance embedding serves as a \textit{steering vector} that modulates the agent’s raw perception $\phi(x_t)$, making it more task-aware. This objective enables the policy to co-adapt to semantic guidance signals, supporting nuanced, language-driven behaviors while maintaining a minimal training budget.
\paragraph{\textsc{Reflector}: Inference-Time Action Refinement}

To enhance robustness during inference, we introduce \textsc{Reflector}, a module that improves action prediction when the VLM exhibits low confidence. At each time step $t$, the VLM, serving as a preliminary decision-maker, receives a multimodal prompt consisting of the current image, task description, and robot state. 
It generates: 1) a step-by-step chain-of-thought (CoT), 2) a textual description of the current condition $C_{str}^t$, and 3) a predicted action instruction $I_{str}^t$. Each prediction is accompanied by a confidence score:
\[
\text{Conf}(a) = \frac{1}{L} \sum_{t=1}^{L} \max \left( \operatorname{softmax}(\mathbf{a}_t) \right),
\]
where $\mathbf{a}_t$ is the logit vector for the $t$-th token in the action sequence of length $L$. This score serves as a practical heuristic for detecting uncertainty. 
If the confidence falls below a predefined threshold $\tau$, the \textsc{Reflector} is triggered.

The \textsc{Reflector} initiates an iterative refinement loop by invoking an LLM to act as a diagnostic engine. It analyzes the VLM's chain-of-thought and task prompt to identify the root cause of the low confidence, such as semantic ambiguity, logical gaps, or failure to account for critical environmental context. Based on this diagnosis, the LLM formulates a targeted query to retrieve the top-$k$ most relevant (condition, action) pairs from prior execution logs. This retrieved knowledge is appended as additional context to a new VLM prompt, which enables the model to re-evaluate and generate a more informed action.

The strength of this framework lies beyond single-instance correction; it establishes a \textit{virtuous cycle of self-improvement}. The final, successfully executed $(C_{str}^t, I_{str}^t)$ pair at each time step is stored in a structured execution log $\mathcal{M}_{task}^k[:t]$. Each time the \textsc{Reflector} resolves an uncertainty, the refined, high-quality action pair is not merely archived—it is integrated into the agent’s memory. This transforms the execution log from a passive record into an active, evolving knowledge base that the system can leverage in future tasks. 
The reflection-driven mechanism thus supports a form of lifelong adaptation, enabling the agent to become progressively more robust and context-aware by accumulating a richer repository of its own successfully navigated experiences.
The complete workflow of this reflection-driven refinement loop is illustrated below:

\begin{mainbox}{Reflector: Iterative Refinement Loop}
\small

\textbf{Trigger:} VLM-generated action has low confidence ($\text{Conf}(a) < \tau$).

\tcbline

\textbf{[Step 1] LLM Reflection \& Query Formulation}
\begin{nestedbox}{\textbf{Input to LLM:}}
    \begin{itemize}
        \item \textbf{VLM's Thinking Steps:} $<$The chain-of-thought trace from VLM$>$
        \item \textbf{Task Description:} $<$The overall goal, e.g., pick up the tomato and place it on the counter$>$
    \end{itemize}
    \textbf{Instruction for LLM:} Given the thinking steps and the task, what key information is still needed to proceed confidently? Output one sentence.
    
    \textbf{Output from LLM (Diagnostic Query):}
    \begin{center}
        \textit{$<$A generated question, e.g., How should I adjust my grasp when the target is partially obstructed by a movable, non-target object?$> $}
    \end{center}
\end{nestedbox}

\tcbline
\textbf{[Step 2] Retrieve Relevant Examples from Execution Log}

\begin{itemize}
    \item \textbf{Retrieved Example 1:} (Condition:  Trying to get mustard, but salt shaker was blocking. Action: Gently nudge the salt shaker to the side, then grasp the mustard.)
    \item \textbf{Retrieved Example 2:} (Condition:  Reaching for a mug, but a book was too close. Action: Tilt the gripper to approach from a sideways angle, avoiding the book.)
\end{itemize}
\tcbline

\textbf{[Step 3] VLM Refinement with Augmented Context}
\begin{nestedbox}{\textbf{New, Augmented Input to VLM:}}
    \begin{itemize}
        \item \textbf{Original Inputs:} $<$Image, Task Description, Robot State$>$
        \item \textbf{Augmented Context (from Step 2):} \\
        Here are some examples to help you decide. Now, re-evaluate and provide the best next action.
    \end{itemize}

\end{nestedbox}
\end{mainbox}


\section{Experiments} \label{sec:exp}
Our extensive experiments answer the following questions: \textbf{Q1)} Can \texttt{GUIDES} improve the success rate of different policies consistently? \textbf{Q2)} How does \textsc{Reflector}'s inference-time reasoning affect the improvement of the success rate? \textbf{Q3)} Is \texttt{GUIDES} effective in real-world tasks?  
\begin{table*}

\centering
\caption{Comparison of w/o G (no GUIDES) vs.\;w/ G (with GUIDES) sorted by decreasing diffusion accuracy improvement on the RoboCasa benchmark.}

\scriptsize
\renewcommand{\arraystretch}{1.1}
\setlength{\tabcolsep}{2pt}
\resizebox{\textwidth}{!}{%
\begin{tabular}{l cc cc l cc cc l cc cc}
\toprule
\textbf{Task} & \multicolumn{2}{c}{\textbf{Transformer}} & \multicolumn{2}{c}{\textbf{Diffusion}} & \textbf{Task} & \multicolumn{2}{c}{\textbf{Transformer}} & \multicolumn{2}{c}{\textbf{Diffusion}} & \textbf{Task} & \multicolumn{2}{c}{\textbf{Transformer}} & \multicolumn{2}{c}{\textbf{Diffusion}} \\
\cmidrule(lr){2-3} \cmidrule(lr){4-5} \cmidrule(lr){7-8} \cmidrule(lr){9-10} \cmidrule(lr){12-13} \cmidrule(lr){14-15}
 & w/o G & w/ G & w/o G & w/ G & & w/o G & w/ G & w/o G & w/ G & & w/o G & w/ G & w/o G & w/ G \\
\midrule
TurnOffSinkFaucet & 0.82 & \textbf{0.84} & 0.12 & \textbf{0.22} & TurnOffMicrowave & 0.80 & \textbf{0.86} & 0.04 & \textbf{0.14} & CloseDrawer & \textbf{1.00} & \textbf{1.00} & 0.20 & \textbf{0.28} \\
TurnOnStove & \textbf{0.56} & \textbf{0.56} & 0.06 & \textbf{0.10} & TurnOffStove & 0.22 & \textbf{0.24} & 0.02 & \textbf{0.04} & TurnSinkSpout & 0.80 & \textbf{0.84} & 0.02 & \textbf{0.04} \\
PnPCabToCounter & 0.22 & \textbf{0.32} & \textbf{0.02} & \textbf{0.02} & PnPCounterToCab & 0.12 & \textbf{0.38} & \textbf{0} & \textbf{0} & PnPCounterToMicrowave & 0.10 & \textbf{0.16} & \textbf{0} & \textbf{0} \\
PnPMicrowaveToCounter & 0.14 & \textbf{0.16} & \textbf{0} & \textbf{0} & PnPSinkToCounter & 0.44 & \textbf{0.60} & \textbf{0} & \textbf{0} & PnPStoveToCounter & 0.36 & \textbf{0.50} & \textbf{0} & \textbf{0} \\
OpenSingleDoor & 0.32 & \textbf{0.54} & \textbf{0} & \textbf{0} & OpenDoubleDoor & 0.18 & \textbf{0.32} & \textbf{0} & \textbf{0} & CloseDoubleDoor & 0.74 & \textbf{0.84} & \textbf{0} & \textbf{0} \\
CloseSingleDoor & 0.88 & \textbf{0.94} & \textbf{0} & \textbf{0} & OpenDrawer & 0.58 & \textbf{0.78} & \textbf{0} & \textbf{0} & CoffeeServeMug & 0.34 & \textbf{0.36} & \textbf{0} & \textbf{0} \\
CoffeeSetupMug & 0.18 & \textbf{0.20} & \textbf{0} & \textbf{0} & CoffeePressButton & 0.64 & \textbf{0.82} & \textbf{0} & \textbf{0} & TurnOnMicrowave & 0.66 & \textbf{0.96} & \textbf{0} & \textbf{0} \\
PnPCounterToSink & 0.26 & \textbf{0.32} & \textbf{0.02} & - & PnPCounterToStove & 0.12 & \textbf{0.14} & \textbf{0.02} & - & TurnOnSinkFaucet & 0.46 & \textbf{0.66} & \textbf{0.02} & - \\
\bottomrule
\end{tabular}%
}
\label{tab:split-benchmark}
\end{table*}
\subsection{Baselines and Experimental Setup}

To validate \texttt{GUIDES}, we conduct experiments in RoboCasa, a high-fidelity simulation benchmark with 24 diverse, long-horizon kitchen tasks. These tasks, ranging from 200 to 700 steps, are challenging due to their complexity and multi-stage structure. 
To demonstrate the versatility of our framework, we evaluate \texttt{GUIDES} using heterogeneous policy architectures, including a behavior cloning transformer and a diffusion model. We adopt DeepSeek-VL-7B-Chat~\cite{lu2024deepseek} as the \textsc{Instructor} to generate semantic guidance, and Qwen2-7B as the \textsc{Reflector} for inference-time reasoning. For real-world validation, we deploy \texttt{GUIDES} on a UR5 robotic platform with an \textsc{Eye-in-Hand} Intel RealSense D435i depth camera~\cite{keselman2017intel}. 
We evaluate the system on the challenging ``PnPCabToCounter'' task, demonstrating its robustness and practicality in physical environments.

\subsection{Experiments in Simulation}

\paragraph{The Performance of \texttt{GUIDES} for Policy Enhancement.}

We evaluate the effectiveness of \texttt{GUIDES} in enhancing pre-trained policies on the RoboCasa benchmark, covering both transformer-based and diffusion-based architectures. In all experiments, we apply our guidance-aware training strategy to inject semantic instructions generated by the \textsc{Instructor}.
To align the pre-trained policy with the newly injected semantic signals, we perform a brief adaptation phase. Our evaluation (Table~\ref{tab:split-benchmark}) starts from a base policy trained for 900 epochs. We compare continued training for an additional 100 epochs (a lightweight adaptation representing only 10\% of the original schedule), with and without \texttt{GUIDES}. For transformer-based policies, \texttt{GUIDES} improves the average absolute success rate by \textit{10 percentage points} and the relative success rate by \textit{33.78\%}. For diffusion-based policies, \texttt{GUIDES} improves performance on tasks with non-zero baseline success by an average of \textit{106\%} in relative terms. 
These results demonstrate that \texttt{GUIDES} serves as a general-purpose performance enhancer across diverse policy architectures, which validates our core hypothesis that supervising a policy on high-level semantic intents instead of low-level kinematics acts as a powerful regularizer, fostering more robust and generalizable behaviors.

\paragraph{Effectiveness of \textsc{reflector}.} \label{exp:infernece}

To evaluate the impact of the \textsc{Reflector}, we test it on three challenging RoboCasa tasks where the base policies perform poorly. Due to the high computational cost of inference-time reasoning, we run 20 trials per task under identical initial conditions to ensure a fair comparison. 
As shown in Table~\ref{tb:reflection}, the \textsc{Reflector} significantly improves performance, achieving a \textit{366.7\% relative improvement} over the baseline and a \textit{47.4\% relative improvement} over \texttt{GUIDES} without reflection.
These results provide strong evidence for our proposed architectural division of labor: many critical task failures are not mere motor execution errors but rather ``strategic stalls'' caused by encountering novel uncertainties. The \textsc{Reflector} introduces a deliberative reasoning loop that enables the agent to resolve such high-level uncertainties online during execution. This problem-solving capability is a powerful complement to the semantic guidance injected into the base policy, which enables the agent to navigate challenging situations with significantly greater robustness.

\paragraph{Gradient Embedding Quality Analysis.}

\begin{figure*}[!tbp]
    \centering
    \includegraphics[width=0.98\textwidth]{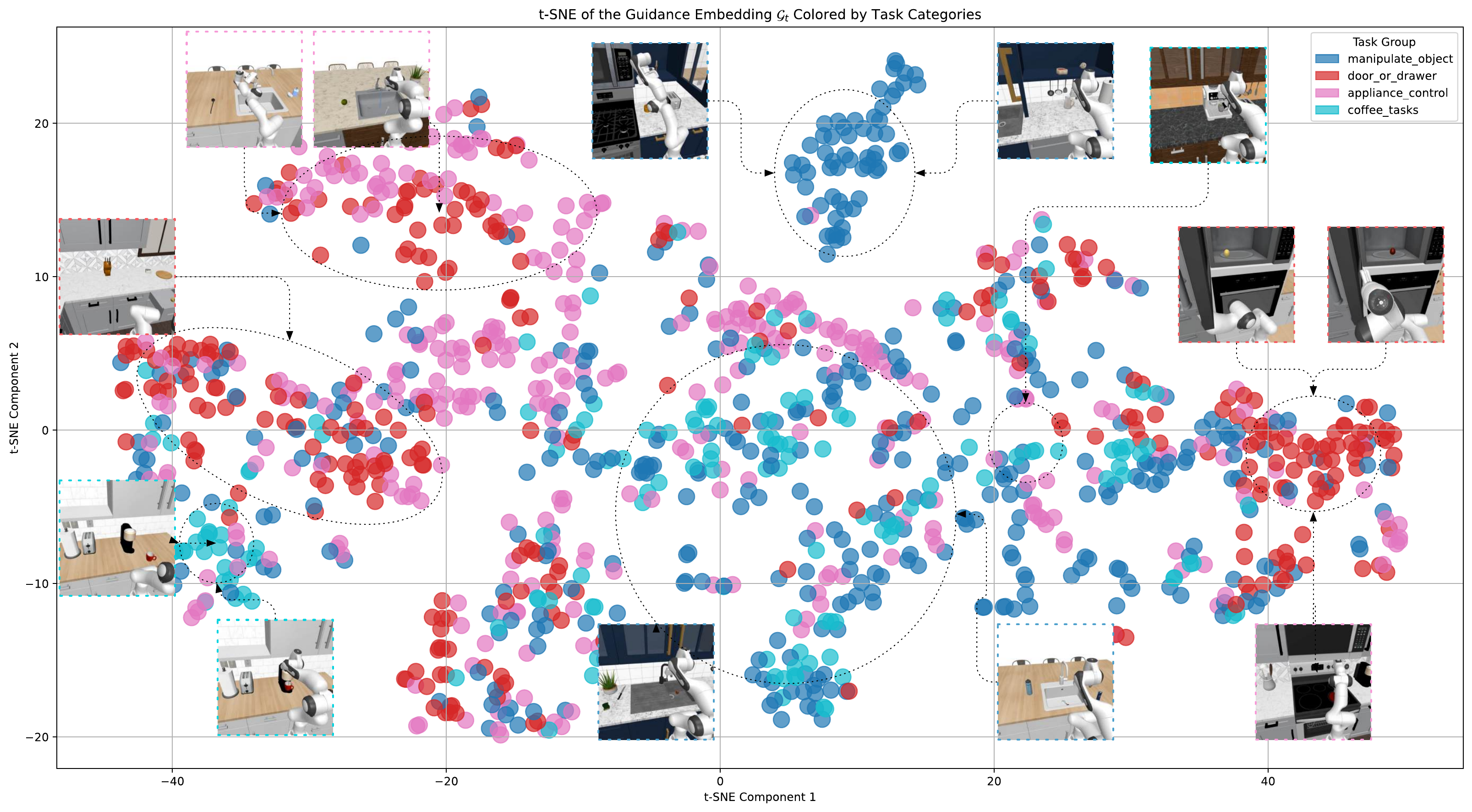}
    \caption{The t-SNE visualization of guidance embeddings ($\mathcal{G}_t$), colored by task category. Note the distinct clusters for manipulation, door/drawer, and appliance-related tasks, which indicate a semantically structured latent space.}
    \vspace{-0.3cm}
    \label{fig:tsen}
\end{figure*}

We use t-distributed stochastic neighbor embedding (t-SNE)~\cite{van2008visualizing} to visualize the guidance embeddings $\mathcal{G}_t$, to examine whether embeddings from different tasks exhibit meaningful semantic structure. The resulting t-SNE plot (Figure~\ref{fig:tsen}) reveals distinct and interpretable clusters. For example, object manipulation tasks form a clear central group, while appliance-control tasks cluster separately toward the right. At a finer level, door and drawer operations subdivide into sub-clusters corresponding to hinged doors and sliding drawers, respectively, which reflect their differing manipulation dynamics. Notably, we observe partial overlap between manipulation and coffee-making tasks, suggesting shared low-level primitives such as grasping and lifting.
This structured organization indicates that the guidance embeddings effectively encode both high-level task categories and fine-grained interaction patterns, which provides semantically rich context to the downstream policy, and remains stable across multiple random seeds.

\subsection{Ablation Study}

We conduct an ablation study to assess the contribution of different components in \texttt{GUIDES} and identify potential performance bottlenecks. Starting from the average success rate of the transformer-based model reported in Table~\ref{tb:ablation}, we evaluate the following modifications:  
1) disabling motion-aware fine-tuning;  
2) removing task descriptions from the AEM input; and  
3) injecting stochastic (i.e., random) values as guidance embeddings $\mathcal{G}_t$ during inference.
The results reveal that $\mathcal{G}_t$ is the most critical component: injecting invalid (random) embeddings leads to complete task failure across all runs, which highlights the necessity of semantically grounded guidance for successful execution. Additionally, motion-aware fine-tuning is essential for producing accurate guidance and substantially impacts overall performance.
Another notable observation from Table~\ref{tb:ablation} is that removing the task description (e.g., ``move the tomato from the cabinet to the counter'') from the AEM input only slightly degrades performance. This suggests that the VLM can generate highly relevant step-wise action instructions even when only the current visual observation is provided. This insight implies the dominant role of visual context in robotic decision making within our framework.

\begin{table} 
\centering
\caption{(a): effect of the \textsc{Reflector} on selected tasks. (b): ablation of key components of GUIDES.}
\label{tab:two_tables_vertical}
\begin{subtable}{\columnwidth} 
\centering\scriptsize
\renewcommand{\arraystretch}{1.1}
\caption{Task-wise performance with \textsc{Reflector}.}
\label{tb:reflection}
\begin{tabular}{@{}lccc@{}}
\toprule
\textbf{Task} & \textbf{BCX} & \textbf{GUIDES} & \textbf{REFLECTOR} \\
\midrule
PnPCounterToCab & 0.12 & 0.38 & 0.56 \\
TurnOffStove                & 0.22 & 0.24 & 0.40 \\
CoffeeSetupMug              & 0.18 & 0.20 & 0.50 \\
\midrule
\textbf{Average}              & 0.17 & 0.27 & 0.49 \\
\bottomrule
\end{tabular}
\end{subtable}

\vspace{0.3cm}
\begin{subtable}{\columnwidth} 
\centering\scriptsize
\renewcommand{\arraystretch}{1.1}
\caption{Ablation study of key GUIDES components.}
\label{tb:ablation} 
\begin{tabular}{lc}
\toprule
\textbf{Method} & \textbf{Success Rate (\%)} \\
\midrule
GUIDES (full)         & 55.1 \\
 w/o motion FT       & 44.5 \\
 w/o task desc.      & 46.0 \\
 w/o efficient AEM & 0.0 \\
\bottomrule
\end{tabular}
\end{subtable}
\end{table}

\subsection{Real Robot Experiment}

To evaluate the generalization capability of \texttt{GUIDES}, we deploy it on a 6-DoF UR5 robotic arm performing the ``PnPCabToCounter'' task. We first generated a dataset of 3,000 trajectories by augmenting 50 expert demonstrations from RoboCasa using MimicGen~\cite{mandlekar2023mimicgen}. Using this dataset, we trained a BC-Transformer visuomotor policy~\cite{mandlekar2021what}, deployed it on the UR5 platform, and measured performance with and without \texttt{GUIDES} integration.

\paragraph{Experimental Setup}  
To ensure the safety of the robot during real-world evaluations, we mounted a \SI{21}{inch}$\times$\SI{24}{inch} cardboard box onto a shelf to serve as an open-front cabinet substitute. An \textsc{Eye-in-Hand} Intel RealSense D435i camera~\cite{keselman2017intel} captures close-range visual observations, providing the input images required by the VLM. During execution, trajectories generated by the trained policy are carried out using the \textsc{scaled-joint-trajectory-controller} provided by the ROS~2 driver.

\begin{figure}[!t]
  \centering
  \includegraphics[width=\linewidth]{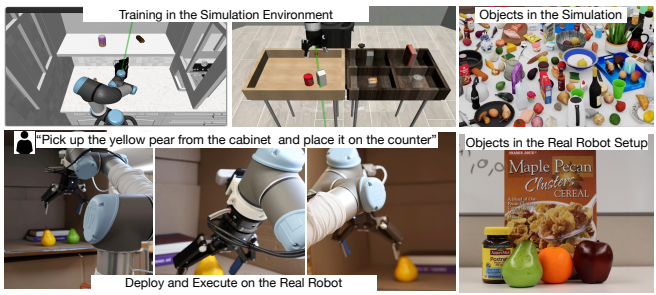}
  \caption{Experimental platform with UR5.}
  \vspace{-0.3cm}
  \label{fig:real_platform}
\end{figure}

\begin{figure}[!tbp]
  \centering
  \includegraphics[width=\linewidth]{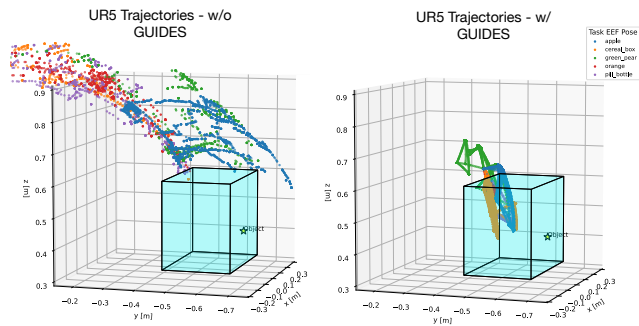}
  \caption{End-effector trajectory distribution (the block denotes the “strike zone”).}
  \label{fig:traj_cmp}
\end{figure}
\paragraph{Control vs.\ Experimental Policies}  
We implement a visuomotor policy based on a BC-Transformer. The \emph{control} policy serves as the baseline, trained entirely in simulation for 1000 epochs without integrating the proposed \texttt{GUIDES} framework. In contrast, the \emph{experimental} policy incorporates \texttt{GUIDES}, branching from the control policy at epoch 900 and undergoing an additional 100 epochs of fine-tuning with guidance integration. 
Crucially, both the \emph{control} and \emph{experimental} policies share the same underlying network architecture, hyperparameters, and initialization before fine-tuning. This controlled setup ensures a direct and fair comparison by holding all other variables constant.

\paragraph{Results and Discussion.}  
To robustly evaluate performance under sim-to-real conditions, we focus on grasp precision, a critical phase in the task, by measuring the frequency with which the end-effector enters a high-precision ``strike zone'' around each object. The impact of \texttt{GUIDES} is substantial (Table~\ref{tab:ur5_inbox_freq}), which increases average strike-zone engagement nearly tenfold, from 4.1\% to \textit{38.8\%}.
This quantitative improvement is supported by qualitative observations (Figure~\ref{fig:traj_cmp}), which show that \texttt{GUIDES} transforms hesitant, drifting trajectories into more decisive and direct motions. While the sim-to-real gap still limits full task success, these results demonstrate that \texttt{GUIDES} effectively translates semantic guidance into significantly more precise and reliable physical behaviors. From a compute perspective, while we do not benchmark wall-clock cost, GUIDES fine-tunes lightweight auxiliaries and gates model calls to uncertain steps, enabling single-GPU development. In contrast, end-to-end VLAs train at scale—for example, OpenVLA \cite{Kim2024} trains on 64×A100 for approximately 2 weeks and needs 15 GB GPU memory for single-GPU inference—so our focus is an upgrade operating point rather than replacement. A full cost–benefit analysis is left for future work.

\begin{table}[!t]
  \centering
  \caption{End-effector ``strike zone'' entry ratio per object}
  \label{tab:ur5_inbox_freq}
  \scriptsize
  \begin{tabular}{lcc}
    \toprule
    \textbf{Object} & \textbf{w/o GUIDES} & \textbf{w/ GUIDES} \\
    \midrule
    apple           & 0.00000             & 0.24792 \\
    cereal box      & 0.00840             & 0.57200 \\
    green pear      & 0.00000             & 0.08288 \\
    orange          & 0.00044             & 0.21383 \\
    pill bottle     & 0.19454             & 0.82058 \\
    \midrule
    \textbf{Overall Mean} & \textbf{0.04068} & \textbf{0.38795} \\
    \bottomrule
  \end{tabular}
\end{table}

\section{Conclusion and Future Work}

We presented \texttt{GUIDES}, a lightweight, architecture-agnostic framework that injects structured guidance from foundation models into pre-trained robotic policies. \texttt{GUIDES} preserves the original policy architecture and embodied knowledge while achieving notable performance gains both in simulation and on a real UR5 arm. These results demonstrate that structured, model-agnostic guidance can substantially enhance policy effectiveness without costly retraining or architectural modifications.
However, \texttt{GUIDES} remains a policy \emph{enhancer}, it cannot bootstrap task success when the baseline policy exhibits no useful behavior. This is because \texttt{GUIDES} functions as a semantic bias that steers the policy within its learned latent manifold, but relies on the base model's pre-trained decoder to execute physically feasible motions. It cannot correct failures where the underlying motor primitives are entirely absent.
Additionally, our experiments exposed a sim-to-real gap, particularly in sensory fidelity.
Future work will: 1) develop more sample-efficient training schedules and knowledge-reuse strategies to address harder tasks; and 2) reduce the sim-to-real gap by incorporating active depth sensing or stereo vision to improve distance estimation and collision avoidance.

\bibliographystyle{IEEEtran}
\bibliography{ieee}

\begin{thebibliography}{10}
\providecommand{\url}[1]{#1}
\csname url@samestyle\endcsname
\providecommand{\newblock}{\relax}
\providecommand{\bibinfo}[2]{#2}
\providecommand{\BIBentrySTDinterwordspacing}{\spaceskip=0pt\relax}
\providecommand{\BIBentryALTinterwordstretchfactor}{4}
\providecommand{\BIBentryALTinterwordspacing}{\spaceskip=\fontdimen2\font plus
\BIBentryALTinterwordstretchfactor\fontdimen3\font minus \fontdimen4\font\relax}
\providecommand{\BIBforeignlanguage}[2]{{%
\expandafter\ifx\csname l@#1\endcsname\relax
\typeout{** WARNING: IEEEtran.bst: No hyphenation pattern has been}%
\typeout{** loaded for the language `#1'. Using the pattern for}%
\typeout{** the default language instead.}%
\else
\language=\csname l@#1\endcsname
\fi
#2}}
\providecommand{\BIBdecl}{\relax}
\BIBdecl

\bibitem{zhang2024vision}
J.~Zhang, J.~Huang, S.~Jin, and S.~Lu, ``Vision-language models for vision tasks: A survey,'' \emph{IEEE Transactions on Pattern Analysis and Machine Intelligence}, 2024.

\bibitem{ghosh2024exploring}
A.~Ghosh, A.~Acharya, S.~Saha, V.~Jain, and A.~Chadha, ``Exploring the frontier of vision-language models: A survey of current methodologies and future directions,'' \emph{arXiv preprint arXiv:2404.07214}, 2024.

\bibitem{wang2024qwen2}
P.~Wang, S.~Bai, S.~Tan, S.~Wang, Z.~Fan, J.~Bai, K.~Chen, X.~Liu, J.~Wang, W.~Ge \emph{et~al.}, ``Qwen2-vl: Enhancing vision-language model's perception of the world at any resolution,'' \emph{arXiv preprint arXiv:2409.12191}, 2024.

\bibitem{team2023gemini}
G.~Team, R.~Anil, S.~Borgeaud, J.-B. Alayrac, J.~Yu, R.~Soricut, J.~Schalkwyk, A.~M. Dai, A.~Hauth, K.~Millican \emph{et~al.}, ``Gemini: a family of highly capable multimodal models,'' \emph{arXiv preprint arXiv:2312.11805}, 2023.

\bibitem{nag2025conformal}
S.~Nag, U.~Ghosh, C.-K. Ta, S.~Bose, J.~Li, and A.~K. Roy-Chowdhury, ``Conformal prediction and mllm aided uncertainty quantification in scene graph generation,'' in \emph{Proceedings of the IEEE/CVF Conference on Computer Vision and Pattern Recognition}, 2025, pp. 11\,676--11\,686.

\bibitem{fan2026vlm}
Z.~Fan, J.~Zhang, R.~Li, J.~Zhang, R.~Chen, H.~Hu, and et~al, ``Vlm-3r: Vision-language models augmented with instruction-aligned 3d reconstruction,'' in \emph{Proceedings of the IEEE/CVF Conference on Computer Vision and Pattern Recognition (CVPR)}, 2026.

\bibitem{touvron2023llama}
H.~Touvron, T.~Lavril, G.~Izacard, X.~Martinet, M.-A. Lachaux, T.~Lacroix, B.~Rozi{\`e}re, N.~Goyal, E.~Hambro, F.~Azhar \emph{et~al.}, ``Llama: Open and efficient foundation language models,'' \emph{arXiv preprint arXiv:2302.13971}, 2023.

\bibitem{naveed2023comprehensive}
H.~Naveed, A.~U. Khan, S.~Qiu, M.~Saqib, S.~Anwar, M.~Usman, N.~Akhtar, N.~Barnes, and A.~Mian, ``A comprehensive overview of large language models,'' \emph{arXiv preprint arXiv:2307.06435}, 2023.

\bibitem{li2023vision}
X.~Li, M.~Liu, H.~Zhang, C.~Yu, J.~Xu, H.~Wu, C.~Cheang, Y.~Jing, W.~Zhang, H.~Liu \emph{et~al.}, ``Vision-language foundation models as effective robot imitators,'' \emph{arXiv preprint arXiv:2311.01378}, 2023.

\bibitem{zawalski2024robotic}
M.~Zawalski, W.~Chen, K.~Pertsch, O.~Mees, C.~Finn, and S.~Levine, ``Robotic control via embodied chain-of-thought reasoning,'' \emph{arXiv preprint arXiv:2407.08693}, 2024.

\bibitem{li2024improving}
J.~Li, Y.~Zhu, Z.~Tang, J.~Wen, M.~Zhu, X.~Liu, C.~Li, R.~Cheng, Y.~Peng, and F.~Feng, ``Improving vision-language-action models via chain-of-affordance,'' \emph{arXiv preprint arXiv:2412.20451}, 2024.

\bibitem{Kim2024}
G.-Y. Kim, J.-H. Shin, J.-H. Bae, C.-H. Kweon, D.-H. Kim, S.-H. Baek, M.-S. Kweon, J.-H. Lee, and J.-H. Kim, ``Openvla: An open-source vision-language-action model,'' \emph{arXiv preprint arXiv:2406.09246}, 2024.

\bibitem{yan2025rdd}
M.~Yan, Y.~Wang, Z.~Liu, and J.~Li, ``Rdd: Retrieval-based demonstration decomposer for planner alignment in long-horizon tasks,'' in \emph{Proceedings of the 39th Annual Conference on Neural Information Processing Systems (NeurIPS)}, 2025.

\bibitem{wang2026drive}
Z.~Wang, H.~Jiang, S.~Dong, Y.~Wang, H.~Qiu, and J.~Li, ``Drive my way: Preference alignment of vision-language-action model for personalized driving,'' in \emph{Proceedings of the IEEE/CVF Conference on Computer Vision and Pattern Recognition (CVPR)}, 2026.

\bibitem{zhang2025lamma}
X.~Zhang, H.~Qin, F.~Wang, Y.~Dong, and J.~Li, ``Lamma-p: Generalizable multi-agent long-horizon task allocation and planning with lm-driven pddl planner,'' in \emph{IEEE International Conference on Robotics and Automation (ICRA)}.\hskip 1em plus 0.5em minus 0.4em\relax IEEE, 2025.

\bibitem{zhang2026commcp}
X.~Zhang, Z.~Wang, Z.~Li, J.~Yao, and J.~Li, ``Commcp: Efficient multi-agent coordination via llm-based communication with conformal prediction,'' in \emph{IEEE International Conference on Robotics and Automation (ICRA)}.\hskip 1em plus 0.5em minus 0.4em\relax IEEE, 2026.

\bibitem{chakraborty2025heal}
T.~Chakraborty, U.~Ghosh, X.~Zhang, F.~F. Niloy, Y.~Dong, J.~Li, A.~Roy-Chowdhury, and C.~Song, ``Heal: An empirical study on hallucinations in embodied agents driven by large language models,'' in \emph{Findings of the Association for Computational Linguistics: EMNLP 2025}, 2025, pp. 21\,226--21\,243.

\bibitem{gong2026autofocus}
L.~Gong, F.~Bahrani, Y.~Zhou, A.~Banayeeanzade, J.~Li, and E.~B{\i}y{\i}k, ``Autofocus-il: Vlm-based saliency maps for data-efficient visual imitation learning without extra human annotations,'' in \emph{IEEE International Conference on Robotics and Automation (ICRA)}, 2026.

\bibitem{kim2025human}
H.~Kim, K.~Lee, J.~Park, J.~Li, and J.~Park, ``Human implicit preference-based policy fine-tuning for multi-agent reinforcement learning in usv swarm,'' in \emph{IEEE/RSJ International Conference on Intelligent Robots and Systems (IROS)}, 2025.

\bibitem{Brohan2023}
A.~Brohan, N.~Brown, J.~Carbajal, Y.~Chebotar, and et~al, ``Rt-2: Vision-language-action models transfer web knowledge to robotic tasks,'' in \emph{Proceedings of the 7th Conference on Robot Learning (CoRL)}, 2023.

\bibitem{mandlekar2021what}
A.~Mandlekar, D.~Xu, J.~Wong, S.~Nasiriany, C.~Wang, R.~Kulkarni, L.~Fei-Fei, S.~Savarese, Y.~Zhu, and R.~Mart{\'\i}n-Mart{\'\i}n, ``What matters in learning from offline human demonstrations for robot manipulation,'' in \emph{Conference on Robot Learning}.\hskip 1em plus 0.5em minus 0.4em\relax PMLR, 2021, pp. 662--673.

\bibitem{Chi2023Diffusion}
C.~Chi, S.~Feng, Y.~Du, Z.~Xu, E.~Cousineau, B.~Burchfiel, and S.~Song, ``Diffusion policy: Visuomotor policy learning via action diffusion,'' in \emph{RSS}, 2023, pp. 27--35.

\bibitem{Nasiriany2024RoboCasa}
S.~Nasiriany, A.~Maddukuri, L.~Zhang, A.~Parikh, A.~Lo, A.~Joshi, A.~Mandlekar, and Y.~Zhu, ``Robocasa: Large-scale simulation of everyday tasks for generalist robots,'' in \emph{CoRL}, 2024.

\bibitem{da2024prompt}
L.~Da, M.~Gao, H.~Mei, and H.~Wei, ``Prompt to transfer: Sim-to-real transfer for traffic signal control with prompt learning,'' in \emph{Proceedings of the AAAI Conference on Artificial Intelligence}, 2024, pp. 82--90.

\bibitem{shridhar2023perceiver}
M.~Shridhar, L.~Manuelli, and D.~Fox, ``Perceiver-actor: A multi-task transformer for robotic manipulation,'' in \emph{Conference on Robot Learning}.\hskip 1em plus 0.5em minus 0.4em\relax PMLR, 2023, pp. 785--799.

\bibitem{Dai2024}
Y.~Dai, J.~Lee, N.~Fazeli, and J.~Chai, ``Racer: Rich language-guided failure recovery policies for imitation learning,'' in \emph{2025 IEEE International Conference on Robotics and Automation (ICRA)}, 2025.

\bibitem{wen2024tinyvla}
J.~Wen, Y.~Zhu, J.~Li, M.~Zhu, and et~al, ``Tinyvla: Toward fast, data-efficient vision-language-action models for robotic manipulation,'' \emph{IEEE Robotics and Automation Letters}, pp. 3988--3995, 2025.

\bibitem{Yang2024VLMTAMP}
Z.~Yang, C.~R. Garrett, D.~Fox, T.~Lozano-P{\'e}rez, and L.~P. Kaelbling, ``Guiding long-horizon task and motion planning with vision language models,'' \emph{arXiv preprint arXiv:2410.02193}, 2024.

\bibitem{torabi2018behavioral}
F.~Torabi, G.~Warnell, and P.~Stone, ``Behavioral cloning from observation,'' \emph{arXiv preprint arXiv:1805.01954}, 2018.

\bibitem{florence2022implicit}
P.~Florence, C.~Lynch, A.~Zeng, O.~A. Ramirez, A.~Wahid, L.~Downs, A.~Wong, J.~Lee, I.~Mordatch, and J.~Tompson, ``Implicit behavioral cloning,'' in \emph{Conference on Robot Learning}, 2022, pp. 158--168.

\bibitem{Pomerleau1989ALVINN}
D.~A. Pomerleau, ``Alvinn: An autonomous land vehicle in a neural network,'' in \emph{NIPS}, 1989, pp. 305--313.

\bibitem{abbeel2004apprenticeship}
P.~Abbeel and A.~Y. Ng, ``Apprenticeship learning via inverse reinforcement learning,'' in \emph{Proceedings of the twenty-first international conference on Machine learning}, 2004, p.~1.

\bibitem{argall2009survey}
B.~D. Argall, S.~Chernova, M.~Veloso, and B.~Browning, ``A survey of robot learning from demonstration,'' \emph{Robotics and Autonomous Systems}, vol.~57, no.~5, pp. 469--483, 2009.

\bibitem{hudson2019gqa}
D.~A. Hudson and C.~D. Manning, ``Gqa: A new dataset for real-world visual reasoning and compositional question answering,'' in \emph{Proceedings of the IEEE/CVF Conference on Computer Vision and Pattern Recognition}, 2019, pp. 6700--6709.

\bibitem{johnson2017clevr}
J.~Johnson, B.~Hariharan, L.~Van Der~Maaten, L.~Fei-Fei, C.~Lawrence~Zitnick, and R.~Girshick, ``Clevr: A diagnostic dataset for compositional language and elementary visual reasoning,'' in \emph{Proceedings of the IEEE Conference on Computer Vision and Pattern Recognition}, 2017, pp. 2901--2910.

\bibitem{anthony2009neural}
M.~Anthony and P.~L. Bartlett, \emph{Neural network learning: Theoretical foundations}.\hskip 1em plus 0.5em minus 0.4em\relax Cambridge University Press, 2009.

\bibitem{lu2024deepseek}
H.~Lu, W.~Liu, B.~Zhang, B.~Wang, K.~Dong, B.~Liu, J.~Sun, T.~Ren, Z.~Li, H.~Yang \emph{et~al.}, ``Deepseek-vl: towards real-world vision-language understanding,'' \emph{arXiv preprint arXiv:2403.05525}, 2024.

\bibitem{keselman2017intel}
L.~Keselman, J.~Iselin~Woodfill, A.~Grunnet-Jepsen, and A.~Bhowmik, ``Intel realsense stereoscopic depth cameras,'' in \emph{Proceedings of the IEEE conference on computer vision and pattern recognition workshops}, 2017, pp. 1--10.

\bibitem{van2008visualizing}
L.~Van~der Maaten and G.~Hinton, ``Visualizing data using t-sne.'' \emph{Journal of Machine Learning Research}, vol.~9, no.~11, 2008.

\bibitem{mandlekar2023mimicgen}
A.~Mandlekar, S.~Nasiriany, B.~Wen, I.~Akinola, Y.~Narang, L.~Fan, Y.~Zhu, and D.~Fox, ``Mimicgen: A data generation system for scalable robot learning using human demonstrations,'' in \emph{Conference on Robot Learning (CoRL)}, 2023.

\end{thebibliography}

\end{document}